\documentclass[10pt,twocolumn,letterpaper]{article}

\usepackage{cvpr}              % To produce the CAMERA-READY version
\usepackage{graphicx}
\usepackage{amsmath}
\usepackage{amssymb}
\usepackage{booktabs}
\usepackage{array}
\usepackage[table]{xcolor}
\usepackage{appendix}
\usepackage[ruled,vlined]{algorithm2e}
\usepackage[table]{xcolor}
\usepackage{xcolor}
\usepackage[symbol]{footmisc}
\definecolor{commentcolor}{RGB}{110,154,155}   % define comment color
\newcommand{\PyComment}[1]{\ttfamily\textcolor{commentcolor}{\# #1}}  % add a "#" before the input text "#1"
\usepackage[pagebackref,breaklinks,colorlinks]{hyperref}

\usepackage[capitalize]{cleveref}
\crefname{section}{Sec.}{Secs.}
\Crefname{section}{Section}{Sections}
\Crefname{table}{Table}{Tables}
\crefname{table}{Tab.}{Tabs.}

\def\cvprPaperID{7316} % *** Enter the CVPR Paper ID here
\def\confName{CVPR}
\def\confYear{2023}

\begin{document}

\title{Task-specific Fine-tuning via Variational Information Bottleneck for Weakly-supervised Pathology Whole Slide Image Classification}

\author{Honglin Li$^{1,2}$, Chenglu Zhu$^{2}$, Yunlong Zhang$^{1,2}$, Yuxuan Sun$^{1,2}$, \\ Zhongyi Shui$^{1,2}$,
 Wenwei Kuang$^{2,3}$, Sunyi Zheng$^{2}$, Lin Yang$^{2}$\thanks{Corresponding author.} \\
$^{1}$College of Computer Science and Technology, Zhejiang University \\  $^{2}$School of Engineering, Westlake University     $^{3}$The University of Hong Kong\\
{\tt\small \{lihonglin,yanglin\}@westlake.edu.cn}
}
% \and
% Second Author \footnote{Corresponding author.}\\
% % First line of institution2 address\\
% % {\tt\small secondauthor@i2.org}
% }
% \author{
% \IEEEauthorblockN{Honglin Li$^{1,2}$, Chenglu Zhu$^{2}$, Yunlong Zhang$^{1,2}$, Yuxuan Sun$^{1,2}$, \\Zhongyi Shui$^{1,2}$, Wenwei Kuang$^{2,3}$, Sunyi Zheng$^{2}$, Lin Yang$^2$\footnote{Corresponding author.}}\\
%     \IEEEauthorblockA{$^1$ College of Computer Science and Technology, Zhejiang University}\\
%     \IEEEauthorblockA{$^2$ School of Engineering, Westlake University}
%     \IEEEauthorblockA{$^3$ The University of Hong Kong}\\
%     {\tt\small \IEEEauthorblockA{\{lihonglin,yanglin\}@westlake.edu.cn}}
%     % \IEEEauthorblockA{}
% }

\maketitle

%%%%%%%%% ABSTRACT
\begin{abstract}
While Multiple Instance Learning (MIL) has shown promising results in digital Pathology Whole Slide Image (WSI) classification, such a paradigm still faces performance problems due to the computational costs on Gigapixel WSIs.
To deal with this problem, most MIL methods utilize a frozen pretrained model from ImageNet to obtain representations first. However, this process may lose essential information owing to the large domain gap and hinder the generalization due to the lack of image-level training-time augmentations.
Though Self-supervised Learning (SSL) proposes viable representation learning schemes, the improvement of the downstream task still needs to be further explored in the conversion from the task-agnostic features of SSL to the task-specifics under the partial label supervised learning.
To alleviate the dilemma of computation cost and performance, we propose an efficient WSI fine-tuning framework motivated by the Information Bottleneck theory. The theory enables the framework to find the minimal sufficient statistics of WSI, thus supporting us to fine-tune the backbone into a task-specific representation only depending on WSI-level weak labels.
The WSI-MIL problem is further analyzed to theoretically deduce our fine-tuning method. Our framework is evaluated on five pathology WSI datasets on various WSI heads. The experimental results of our fine-tuned representations show significant improvements in both accuracy and generalization compared with previous works.
Source code will be available at \href{https://github.com/invoker-LL/WSI-finetuning}{https://github.com/invoker-LL/WSI-finetuning}. 
\end{abstract}
%%%%%%%%% BODY TEXT
\begin{figure}[htbp]
  \centering
%  \fbox{\rule{0pt}{2in} \rule{0.9\linewidth}{0pt}}
   \includegraphics[width=1.0\linewidth]{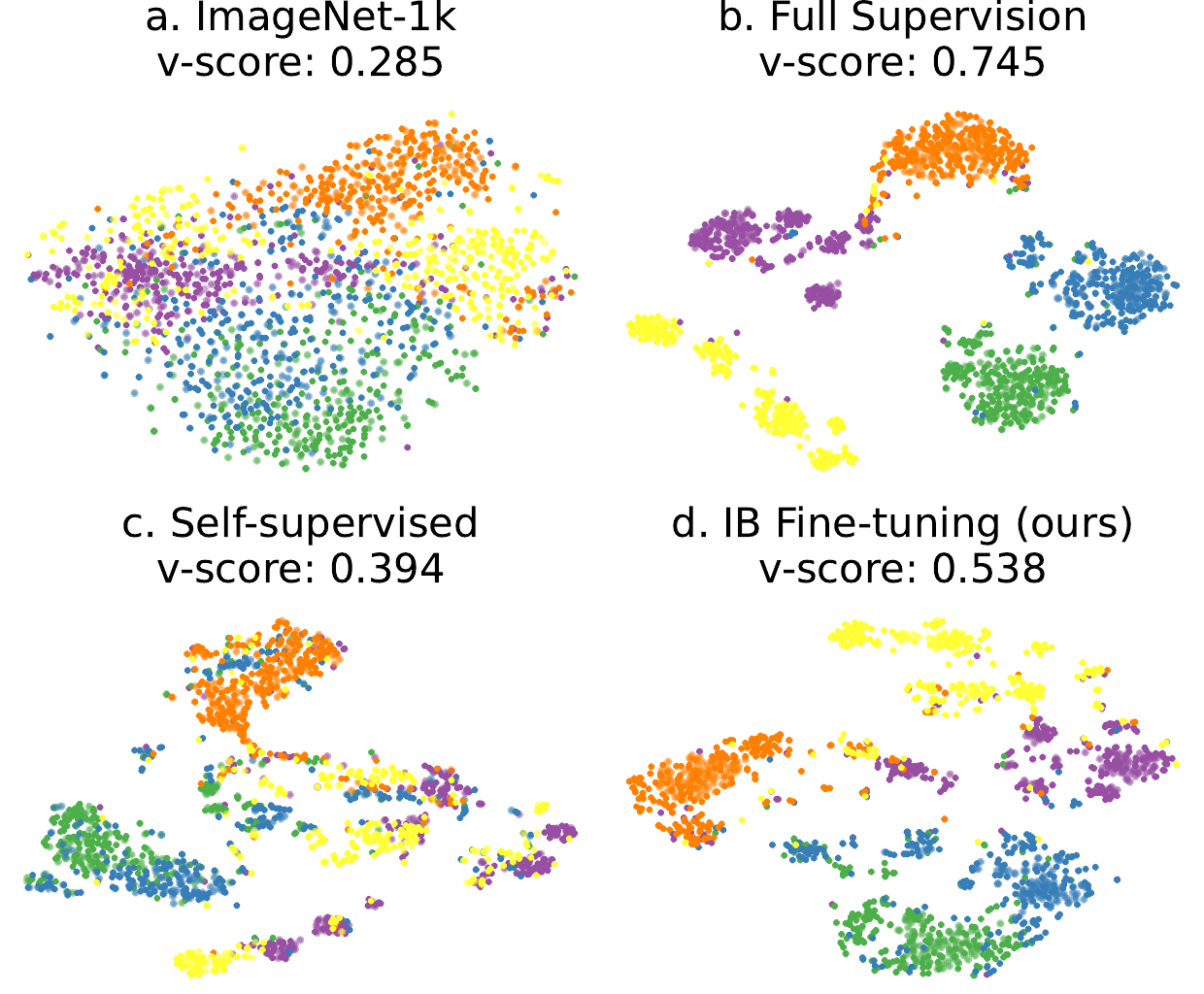}

   \caption{T-SNE visualization of different representations on patches. Our method converts chaotic ImageNet-1K and SSL features into a more task-specific and separable distribution. The cluster evaluation measurement, v-scores, show weakly supervised fine-tuned features are more close to full supervision compared to others. a. ImageNet-1k pretraining. b. Full patch supervision. 
   c. Self-supervised Learning. d. Fine-tuning with WSI labels.}
   \vspace{-3 mm}
   \label{FIG1}
\end{figure}

\section{Introduction}
\label{sec:intro}
Digital pathology or microscopic images have been widely used for the diagnosis of cancers such as Breast Cancer \cite{10.1001/jama.2017.14585} and Prostate Cancer \cite{bulten2022artificial}. However, 
% the diagnosis process needs many efforts of pathologists to scan all views of a Whole Slide Image (WSI), 
the reading of Whole Slide Images (WSIs) with gigapixel resolution is time-consuming which poses an urgent need for automatic computer-assisted diagnosis.
Though computers can boost the speed of the diagnosis process, the enormous size of resolution, over 100M\cite{Zhang2022DTFDMILDF}, makes it infeasible to acquire precise and exhaustive annotations for model training, let alone the current hardware can hardly support the parallel training on all patches of a WSI.
Hence, an annotation-efficient learning scheme with light computation is increasingly desirable to cope with those problems.
In pathology WSI analysis, the heavy annotation cost is usually alleviated by Multiple Instance Learning (MIL) with only WSI-level weak supervision,
which makes a comprehensive decision on a series of instances as a bag sample \cite{maron1997framework,pmlr-v80-ilse18a}.
Intuitively, all small patches in the WSI are regarded as instances to constitute a bag sample \cite{campanella2019clinical,lu2021data, NEURIPS2021_10c272d0}, where the WSI's category corresponds to the max lesion level of all patch instances.

However, most methods pay much effort to design WSI architectures while overlooking the instance-level representation ability. Because of the computational limitation, the gradient at the WSI-level is impossible to parallelly backpropagate to instance encoders with more than 10k instances of a bag. Thus parameters of the pretrained backbone from ImageNet-1k (IN-1K) are frozen to obtain invariant embeddings.
Due to the large domain gap between IN-1K and pathological images, some essential information may be discarded by layers of frozen convolutional filters, which constrains the accuracy of previous WSI methods.
% upper bound for accuracy of WSI classification.
To address the constraint, recent works \cite{li2021dual, chen2022scaling} make efforts to learn a good feature representation at the patch-level by leveraging Self-supervised Learning (SSL).
However, such task-agnostic features are dominated by the proxy objective of SSL, e.g. Contrastive Learning in \cite{MOCO, dino,chen2020simple} may push away the distance between two instances within the same category, thus only performs slightly better than IN-1K pretraining in WSI classification. Nearly all SSL methods \cite{MOCO, dino,chen2020simple, MAE} proposed on natural image recognition utilize a small portion of annotations to get promising fine-tuning accuracy compared to full supervision, which is higher than Linear Probing \cite{MOCO} by a large margin. 

These findings illuminate us to design a fine-tuning scheme for WSI analysis to convert IN-K or SSL task-agnostic representations into task-specifics. Motivated by the Information Bottleneck (IB) theory \cite{alemi2017deep,achille2018information}, we argue that pretraining is limited to downstream tasks, therefore fine-tuning is necessary for WSI analysis. In addition, we develop a solution based on Variational IB to tackle the dilemma of fine-tuning and computational limitation by its minimal sufficient statistics and attribution properties \cite{achille2018information, lehmann2012completeness}. The differences among the above three feature representations are depicted in Figure \ref{FIG1}, where the feature representation under full patch-level supervision is considered as the upper bound.

Our main contributions are in 3 folds: 1) We propose a simple agent task of WSI-MIL by introducing an IB module that distills over 10k redundant instances within a bag into less than 1k of the most supported instances. Thus the parallel computation cost of gradient-based training on Gigapixel Images is over ten times relieved. 
By learning and making classification on the simplified bag, we find that there are trivial information losses due to the low-rank property of pathological WSI, and the distilled bag makes it possible to train a WSI-MIL model with the feature extractor on patches end-to-end, thus boosting the final performance.
2) We argue that the performance can be further improved by combining with the SSL pretraining since we could convert the task-agnostic representation from SSL into task-specific one by well-designed fine-tuning.
The proposed framework only relies on annotations at WSI levels, which is similar to recent SSL approaches \cite{MOCO,chen2020simple,dino,MAE}. Note that our method only utilizes less than a 1\% fraction of full patch annotation to achieve competitive accuracy compared to counterparts.
3) Versatile training-time augmentations can be incorporated with our proposed fine-tuning scheme,
thus resulting in better generalization in various real-world or simulated datasets with domain shift, which previous works ignore to validate.
These empirical results show that our method advances accuracy and generalization simultaneously, and thus would be more practical for real-world applications.

%-------------------------------------------------------------------------
\section{Related Work}
\subsection{Multiple Instance Learning for WSI Analysis}
Multiple Instance Learning (MIL) is a well-defined task and has been explored extensively.
Currently, there are mainly two lines of methods for WSI analysis:
 1) explicit modeling the MIL definition $Y= pooling \{ y_1,y_2,...,y_n \} $ that WSI level prediction is aggregated by the probability of all patches with Mean or Max-pooling \cite{campanella2019clinical,Zhang2022DTFDMILDF}.
 2) implicit learning WSI level representation by aggregating all patches embeddings via a WSI classifier with Recurrent Neural Network (RNN) \cite{campanella2019clinical} or in an attention\cite{pmlr-v80-ilse18a} mechanism.
  The latter shows superior performance since such modeling includes less inductive bias compared to the former processing with fixed weights.

RNN \cite{campanella2019clinical}
  treats WSI as a sequence of patches, but its permutation variation and defects on long-term dependency do not match the nature of WSI well.
By contrast, attention-based MIL (AB-MIL)\cite{pmlr-v80-ilse18a} learns the weights of instance representations adaptively, and these methods can be flexibly inserted into the classification framework as a plug-and-play module to generalize the full-field map according to the task.
 To make patches' representation more discriminative,
 % patches' representation -> patch representation
%  CLAM \cite{lu2021data} proposed to use high attention weights as a positive pseudo label and low attention weights as negative, a patch classifier is trained parallel to WSI classifier.
 CLAM \cite{lu2021data} introduces an auxiliary task in the MIL framework to distinguish its corresponding availability according to the size of instance attention while training the WSI classification.
 Trans-MIL \cite{NEURIPS2021_10c272d0} is designed to model the relation among patches via Self-Attention \cite{NIPS2017_3f5ee243} and solves the softmax computational complexity in the long sequence by linear Self-Attention \cite{wang2020linformer,xiong2021nystromformer}.
 To embed the multi-scale information of WSI, DS-MIL \cite{li2021dual} simply concatenates patch features in three scales, and HIPT \cite{chen2022scaling} builds hierarchical multi-scale features.
To mitigate overfitting in complex attention model trained on the limited number of WSIs, DTFD-MIL \cite{Zhang2022DTFDMILDF} resamples instances from the original bag to generate various sub-bags for augmentation. However, all these works pay too much attention on WSI head to explore the backbone.
\begin{figure*}[hbtp]
  \begin{center}
  \includegraphics[width=0.9\textwidth]{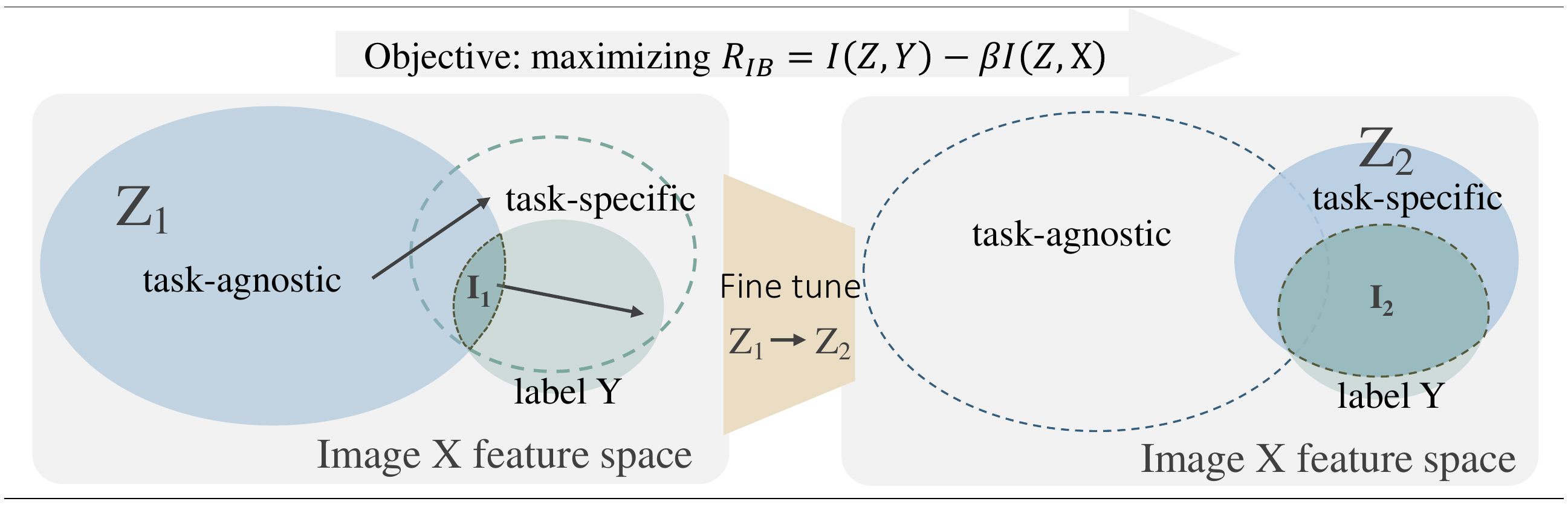}
  \end{center}
  \vspace*{-3 mm}
   \caption{An illustration of the tuning scheme. To maximize $R_{IB}$, enlarging the first term helps latent representation vector Z become more predictive for label Y (More overlap between Y and Z2 compared to Z1, which denotes higher mutual information), while by reducing the second term to filter irrelevant features (smaller overlap between image X and Z2 compared to Z1).}
    \label{FIG2}
  \vspace{-2 mm}
  \end{figure*}
\subsection{Self-supervised Learning and Fine-tuning}
Self-supervised Learning (SSL) has shown to be a promising paradigm both in computer vision \cite{MOCO,chen2020simple,MAE,dino,DBLP:journals/corr/abs-2106-08254} and natural language processing \cite{peters-etal-2018-deep,devlin2018pretraining,Yang2019XLNetGA}.
However, fine-tuning (FT) is quite necessary for the pretrained model of SSL to the downstream tasks,
e.g. in \cite{MAE,MOCO}, the vanilla FT method can improve about 15 percent of accuracy in IN-1K compared to vanilla Linear Probing\cite{MOCO}. More interesting and parameter-efficient FT methods \cite{floridi2020gpt,li2021prefix,lester2021power,hu2021lora,li2020few,zhao2022few,Zhao_2022_CVPR} are proposed according to the similarity between pretraining and downstream in natural language processing and few-shot image generation.
Unfortunately, the computational limitation on redundant instances prevents FT from being directly introduced into WSI analysis. Although SSL achieves optimistic performance in some works \cite{li2021dual,chen2022scaling} for pathology WSI analysis but only shows a small margin of improvement. There may be much potential to be discovered in WSI analysis if an appropriate FT method can be used to address the training cost from surplus information of WSI.

\subsection{Information Bottleneck and Attributions}
The Information Bottleneck (IB) conception is introduced in \cite{tishby2000information} as an information-theoretic framework for learning, which is currently used in deep learning both theoretically and practically.
\cite{shwartz2017opening} proposes to unveil deep neural network in an information flow perspective by estimating Mutual Information (MI) between the outputs of two layers, then in followed-up works \cite{saxe2019information,goldfeld2018estimating}, they revisit the MI compression process of IB to propose complements and better measurements of MI in Deep Neural Network (DNN).

IB not only reveals how the traditional DNN training finishes information compression implicitly in the above works, but can also be employed explicitly as an objective to intervene in DNN's training: \cite{alemi2017deep} derives a variational bound of the IB objective, claiming that the proxy objective provides an additional regularization term which improves the robustness to adversarial samples compared to the traditional training of DNN.
\cite{paranjape-etal-2020-information} adopts
the variational bound in \cite{alemi2017deep} and replaces the distribution of latent features from Gaussian into Bernoulli. In this way, it generates masks to control the conciseness of the rationale extraction of sentences.
Achille et al.\cite{achille2018information} claim that IB solutions approximate minimal sufficient statistics (MSS) \cite{lehmann2012completeness}, and by penalizing the redundancy of representations, they heuristically argue that the model’s sensitivity to nuisances is mitigated.
Similar to some visual attribution methods like CAM \cite{CAM,Grad-CAM}, LRP\cite{bach2015pixel} and patch masking \cite{MAE,DBLP:journals/corr/abs-2106-08254,liang2022not} in ViT, an IB-based attribution method is proposed in
\cite{schulz2020restricting} by adding noise to intermediate feature maps, restricting the flow of information, then how much information image regions provide can be quantified.
Intuitively, the IB objective reveals the limitation of pretraining for downstream tasks and the imperative of FT for task-specific representation in WSI analysis, as shown in Figure \ref{FIG2}. Moreover, it motivates us to solve the dilemma of FT and computational limitation by its' MSS and visual attribution properties.

  \section{Method}

  \subsection{Overview on MIL-based WSI Analysis}
  
  % \textbf{WSI classification with Slide-Level Weak Supervision}:
  Given a WSI $X$, the goal is to make slide-level prediction $Y$ by learning a classifier $f(X;\theta)$.
  Due to its extremely high resolution, X is patched into a huge bag of small instances $X=\{x_1,..., x_N\}$, where N is the number of instance.
  The slide-level supervision $\hat{Y}$ is given by a Max-pooling operation of the latent label $\hat{y_i}$ for each instance $x_i$, which can be defined as:
  \begin{equation}
    \hat{Y}=\max\{\hat{y_1},..., \hat{y}_N\}.
    \label{EQ1}
  \end{equation}
  Since all latent labels of instances $\hat{y_i}$ are unknown under the WSI-level supervision,
  conventional approaches convert this problem into a MIL formulation in the following two steps:
  % it is usually converted into a MIL formulation as following two steps:
  1) Processing images into feature representations $Z=\{z_1,...,z_N\}$ with a backbone $h$ as $z_i=h(x_i;\theta_1)$ where $h$ is a model of any architecture such as CNN or ViT with parameters $\theta_1$.
  2) Aggregating all patches' features within a slide and producing the slide-level prediction $Y = g(Z; \theta_2)$,
   where $g$ is an attention-based pooling function followed by a linear classifier head as:
   % 这里不用e.g. 不是举例， 直接用 as
  \begin{equation}
      g(Z; \theta_2)= \sigma (\sum_{i=1}^N a_i z_i),
      \label{EQ2}
  \end{equation}
   where $a_i$ is attention weights and $\sigma(\cdot)$ is a linear head.
  Limited by the computational cost,
  the parameters $\theta_1$ and $\theta_2$ in $f(X;\theta) = g\{h(X; \theta_1); \theta_2\}$ are learned separately by following steps: 
   1) 
  Initializing $\theta_1$ from the pretrained model, which refers to general features from public IN-1K, or learned by SSL on the related dataset to extract the domain-specific representations.
   2) 
   %freeze $\theta_1$, then update $\theta_2$.
   Freezing $\theta_1$ and learning $\theta_2$ under slide-level supervision.
    
  \subsection{Information Bottleneck for MIL Sparsity}
  % Variational Bound for the IB

  \noindent \textbf{Background of the Information Bottleneck}

  The Information Bottleneck (IB) can work as an information compression role to intervene in DNN's training \cite{alemi2017deep}.
  The objective function of IB to be maximized is given in \cite{tishby2000information} as,
  %\begin{center}
  \begin{equation}
  %  \max \{ \underbrace{I(Z, Y)}_{task~specifics} - \overbrace{ \beta I(Z, X)}^{task~irrelevents} \}
  R_{IB} =  I(Z, Y)- \beta I(Z, X),
  \label{EQ3}
   \end{equation}
  where $I(\cdot,\cdot)$ indicates the Mutual Information (MI)
  and $\beta$ is a Lagrange multiplier controlling the trade-off between the
  information that the representation variable $Z$ shares with the label $Y$ and its shares with input $X$.
  % and its shares with $X$
  Since the computation of MI is intractable during the training of the neural networks,
  to maximize IB objective can be transferred to minimize a variational bound of Eq.\eqref{EQ3} derived in \cite{alemi2017deep} follows:
  % the variational formulation \cite{alemi2017deep can be adopted to convert the maximization of $R_{IB}$ to the minimization of a variational bound as followed,
  \begin{equation}
    \begin{aligned}
  J_{IB} = \frac{1}{N}\sum_{n=1}^N  \mathbb{E}_{z \sim p_{\theta}(z|x_n)}
        [- \log q_{\phi}(y_n|z)] +
    \\ \beta KL[p_{\theta}(z|x_n),r(z)],
    \end{aligned}
    \label{EQ4}
  \end{equation}
  % where $n, N$ denote the sample's index and quantity respectively;
  where $N$ denotes the number of samples,
  $q_{\phi}(y|z)$ is a parametric approximation to the likelihood $p(y|z)$,
  %用逗号不用分号
  $r(z)$ is the prior probability of $z$ to variational approximate the marginal $p(z)$, and $p_{\theta}(z|x)$ is the parametric posterior distribution over $z$.
  %需要起个头

    \begin{figure*}

      \begin{center}
      \includegraphics[width=1.0\textwidth]{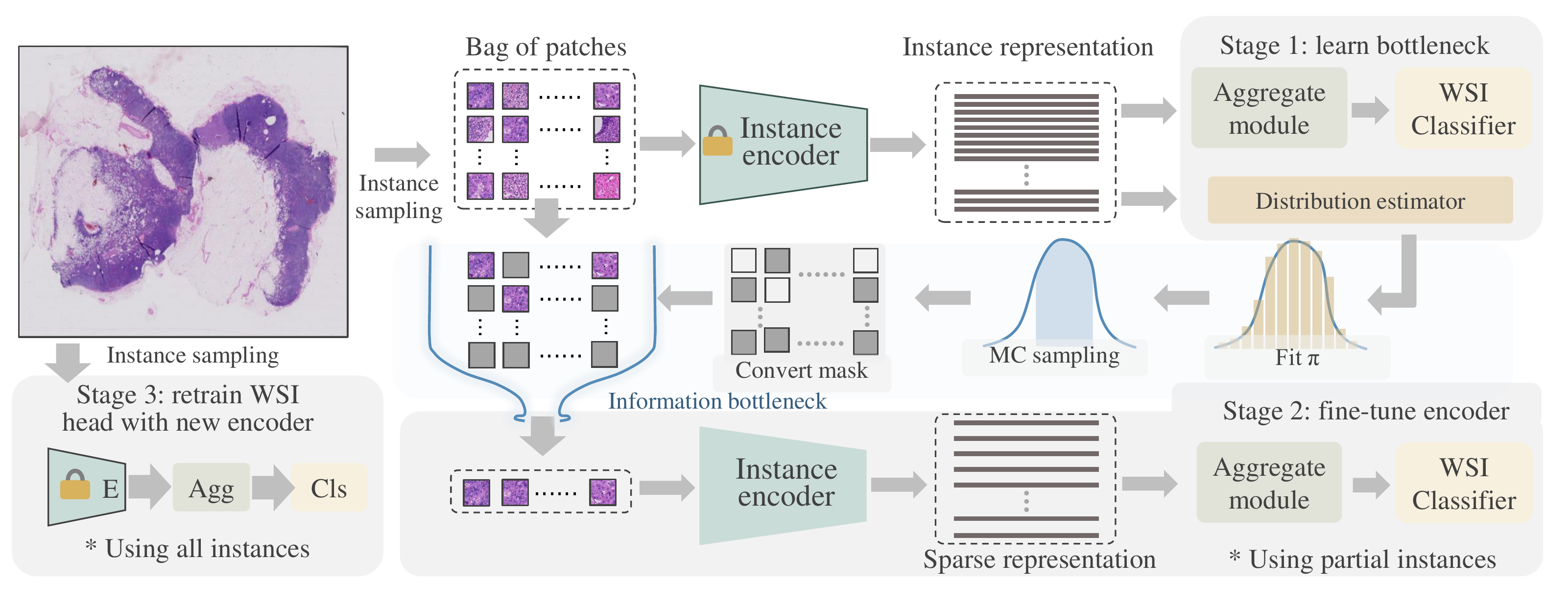}
      \end{center}
      \vspace*{-4 mm}
       \caption{Workflow of WSI-MIL task-specific fine-tuning. 1) Initialize the backbone with pretrained parameters and set frozen, then learn the IB module to generate instance masks. 2) fix the mask to distill a sparse bag, then fine-tune the WSI head and patch the backbone end-2-end. 3) utilize all fine-tuned instance features within a bag and train the WSI-MIL classifier head.}
        \label{FIG3}
      \vspace*{-3 mm}
      \end{figure*}

  \noindent \textbf{Learn MIL Sparsity via Variational Bound}

  To trade off the dilemma of computational limitation and task-specific representation learning via end-to-end back-propagation, we propose to utilize the IB module to filter most task-irrelevant instances for task-specific fine-tuning.

  The above filtering process can be implemented by optimizing the second term of in Eq.\eqref{EQ3} which controls the compression. There are two ways that compress $X$ to $Z$ by decreasing the KL divergence between $p(z|x)$ and $r(z)$ in Eq.\eqref{EQ4} variational method: reducing the dimension of representation $Z$ compared to $X$ in \cite{alemi2017deep}, or converting input $X$ into a sparse one in \cite{paranjape-etal-2020-information}.

  For the setting of our long instance sequenced MIL, we reduce $I(X, Z)$ into a degree so that the gradients can be back-propagated to the backbone encoder, which needs us to convert a WSI of bag size over 10k into 1k for the sake of sparsity. Considering MIL for tumor v.s. normal binary classification without loss of generality and the latent label ${y_i}$ of each instance $x_i$ in Eq.\eqref{EQ1}, we argue that it is sufficient enough to make the WSI level prediction if one tumor area is detected.
  % More generally, for a tumor-positive WSI, if there is at least one tumor patch in a small portion area of WSI, then the small portion area is sufficient enough to make WSI level prediction. Similar for a tumor-negative WSI, if the most suspective area is also negative, there is high a confidence to make negative WSI diagnosis.
  With the above understanding, we propose to learn compressed components similar to \cite{paranjape-etal-2020-information} by defining a IB module as:
  \begin{equation}
  z = m \odot x,
  \label{EQ5}
  \end{equation}
  where $m$ is a Bernoulli$(\pi)$ distributed binary mask and in this way $KL[p_{\theta}(z|x),r(z)]$ in Eq.\eqref{EQ4} can be decomposed as,
  \begin{equation}
      KL[p_{\theta}(m_i|x),r(m_i)] + \pi H(X),
      \label{EQ6}
  \end{equation}
  % where $H(X)$ is the entropy of $X$,
  % thus it can be dropped since it is constant.
  where $H(X)$ is the entropy of $X$, which can be omitted during the minimization due to its constant value. Please check Supplementary for above proof.
  
  The Bernoulli$(\pi)$ distribution for $m$ fits the definition of MIL empirically: 
  we can treat $m$ as a latent weak prediction $\hat{y}$ describing whether the patch contains tumor or not, denoting $P_{set} = \{p(m_1|x_1),..., p(m_N|x_N)\}$, then during inference Eq.\eqref{EQ1} can be derived as:
  % thus the Eq.\eqref{EQ1} can be derived during prediction inference as followed,
  \begin{equation}
    % \begin{aligned}
    \hat{Y} = \max \{P_{set} \} = \max \{P_{subset} \} ,
    % \end{aligned}
    \label{EQ7}
  \end{equation}
  where $P_{subset} \in P_{set}$, generated by select top-K elements in $P_{set}$.
  The patch classifier trained with only slide-level supervision shows low accuracy \cite{li2021dual}, so we only use it to generate mask for sparse sub-bag and still utilize attention-based MIL on the sub-bag for decision making.
  
  \subsection{Loss Function and its Implementation}
  Derived from Eq.\eqref{EQ4} and Eq.\eqref{EQ6}, our Variational IB module should be optimized with the following loss function,
  \begin{equation}
    \begin{aligned}
  loss = \frac{1}{N}\sum_{n=1}^N  \mathbb{E}_{z \sim p_{\theta}(z|x_n)}
        [- \log q_{\phi}(y_n|z)] +
    \\ \beta KL[p_{\theta}(m|x_n),r(m)],
    \end{aligned}
    \label{EQ8}
  \end{equation}
  where the first term is the task loss to learn task-specific features, and it can be treated as the cross entropy same to prior works\cite{paranjape-etal-2020-information, alemi2017deep} in a sampling perspective.
  The second term is the information loss to filter out task-irrelevant instances by minimizing the Kullback-Leibler divergence between the distribution of mask and the prior Bernoulli.
  
  The $p(m_n|x_n)$ is first generated by a linear layer from the representation of $x_n$ and followed by a sigmoid activation. If the mask $m$ is directly converted by thresholding $p(m_n|x_n)$, the gradients from $z = m \odot x$ can not be backpropagated since such an operation is not differentiable.
  So, during training, $m$ is generated via Monte Carlo sampling, and $p(m_n|x_n)$ can be learned with the reparameterization trick \cite{kingma2013auto,kingma2015variational} for gradient's estimation.
  % In the inference stage, to meet Deep Learning frameworks' property that the tensor shape should be the same within a batch, top-K instances are sampled by ranking $p(m_j|x_j)$ within a WSI.
  %In the inference, top-K instances are selected after ranking all instance predictions extracted from a WSI, which satisfies the consistency of tensor shape within a batch in the deep learning framework。（which从句不写也可以，如果放到实现细节的部分，感觉写了有点太啰嗦）
  
  \subsection{Task-specific Fine-tuning}
  
  % To meet the computational limitation, the parameter $\pi$ of Bernoulli distribution $r(m)$ controls the sparsity can be set to $0.025 \sim 0.1$, resulting in a $256 \sim 1024$ bag size for example if the original bag size approximately equals to 10k.
  
  To realize task-specific fine-tuning and better performance with only slide-level labels, the workflow of our method is depicted in Figure \ref{FIG3}, which includes 3 stages:
  1) By minimizing the loss function in section 3.3 with a frozen backbone, a sparse instance set within a bag is obtained.
  2) Assisted by the small portion of instances in a bag generated from stage 1, it is possible to fine-tune the backbone for better task-specific features and here we directly backpropagate the loss gradients from slide-level supervision into top-K patch instances.
  3) Since the appropriateness of the sparse instance set heavily relies on the performance of the pretrained backbone, the first training IB module is insufficient and may result in small or even zero top-K recall.
  Moreover, the sparse instance set may lose the contextual dependency of WSI modeling.
  Thus in this stage, We utilize all fine-tuned instance features within a bag to train a traditional attention-based WSI-MIL classifier.
  
  % Specifically, we Thus all modules should be optimized iteratively in the proposed framework. 

% \begin{algorithm}[h]
%   \SetAlgoLined
%   \PyComment{Learn sparsity of WSI with fixed backbone} \\
%   for (X,y) in loader:\\
%   \Indp   % start indent
%     with torch.no\_grad():\\
%   \Indp   % start indent
%       Z = model(X)\\
%   \Indm
%       M = IB(Z)\\
%       Y = model\_wsi(M . Z)\\
%     logits = torch.logsigmoid(FC(Z))\\
%     q\_z = RelaxedBernoulli(logits)\\
%     Z\_mask = q\_z.sample()\\
%     p\_z = Bernoulli(probs=p\_prior)\\
    
%     loss1 = CrossEntropyLoss(Y,y)\\
%     loss2 = KL\_divergence(q\_z, p\_z)\\

%   \caption{PyTorch-style pseudocode for WSI Task-specific fine-tuning via VIB}
%   \label{alg}
%  \end{algorithm}

\section{Experiments}
In this section, we present the performance of the proposed method incorporated with the latest WSI-MIL frameworks, where the IN-1K and SSL features are used for comparisons. Ablation experiments are performed to further study the proposed method and for paper length, more experimental results are presented in the Supplementary.

\noindent \textbf{Datasets and Tasks.}

We use five datasets to evaluate our method.
The slide-level classification performance of IN-1K, SSL, and our method is evaluated on three datasets including both histopathology and cytopathology images: two public histopathological WSI datasets, Camelyon-16 \cite{bejnordi2017diagnostic} for tumor / normal binary classification, The Cancer Genome Atlas Breast Cancer (TCGA-BRCA) \cite{petrick2021spie} for tumors subtyping. One internal cytopathology WSI dataset is introduced to validate the universality of our method on both histo- and cyto-pathology, which is Liquid-based Preparation cytology for Cervical Cancer's early screening (LBP-CECA). 

Despite the original evaluation on the closed dataset, we also 
%closed??
evaluate the generalization of our method on Camelyon-16-C (generated with random synthetic domain shift from Camelyon-16), and  Camelyon-17\cite{litjens20181399} from five different centers, which occurs frequently in practical pathological diagnosis and has been hindering the application of automatical WSI analysis to the real world. 
Details of five datasets can be found in the supplementary.

For pre-processing, we follow the operations in CLAM-SB\cite{lu2021data} which mainly includes HSV, Blur, Thresholding, and Contours methods to localize the tissue regions in each WSI. Then non-overlapping patches with size 256 × 256 on the 20X magnification are extracted from the tissue regions. 
\begin{table*}[t]
  \begin{center}   
  % \begin{tabular}{*{7}{c}}
  \begin{tabular}{m{2.7cm}<{}||m{1.9cm}<{\centering}m{1.9cm}<{\centering}|m{1.9cm}<{\centering} m{1.9cm}<{\centering}|m{1.9cm}<{\centering}m{1.9cm}<{\centering}}
%   \begin{tabular}{@{}lc@{}}
  
    \midrule[1.2pt]
    
      & \multicolumn{2}{c|}{\underline{Camelyon-16}} & \multicolumn{2}{c|}{\underline{TCGA-BRCA}}  & \multicolumn{2}{c}{\underline{LBP-CECA}} \\
    Method & F1 & AUC & F1 & AUC & F1 & AUC \\
  
  \midrule
  Full Supervision &  0.967\footnotesize{$\pm$}0.005 & 0.992\footnotesize{$\pm$}0.003  & -&-  & 0.741\footnotesize{$\pm$}0.006 & 0.942\footnotesize{$\pm$}0.002\\
  RNN-MIL \cite{campanella2019clinical} &  0.834\footnotesize{$\pm$}0.017 & 0.861\footnotesize{$\pm$}0.021  & 0.776\footnotesize{$\pm$}0.035 & 0.871\footnotesize{$\pm$}0.033  & - & -\\
  AB-MIL \cite{pmlr-v80-ilse18a} &  0.828\footnotesize{$\pm$}0.013 & 0.851\footnotesize{$\pm$}0.025  & 0.771\footnotesize{$\pm$}0.040 & 0.869\footnotesize{$\pm$}0.037  &  0.525\footnotesize{$\pm$}0.017 & 0.845\footnotesize{$\pm$}0.002 \\
  DS-MIL\cite{li2021dual} &  0.857\footnotesize{$\pm$}0.023 & 0.892\footnotesize{$\pm$}0.012  &  0.775\footnotesize{$\pm$}0.044 & 0.875\footnotesize{$\pm$}0.041  & - & -\\
  % \midrule
  CLAM-SB \cite{lu2021data} & 0.839\footnotesize{$\pm$}0.018 & 0.875\footnotesize{$\pm$}0.028 &  0.797\footnotesize{$\pm$}0.046 & 0.879\footnotesize{$\pm$}0.019 &  0.587\footnotesize{$\pm$}0.014 & 0.860\footnotesize{$\pm$}0.005 \\
  TransMIL \cite{NEURIPS2021_10c272d0} &  0.846\footnotesize{$\pm$}0.013 & 0.883\footnotesize{$\pm$}0.009 & 0.806\footnotesize{$\pm$}0.046 & 0.889\footnotesize{$\pm$}0.036 &  0.533\footnotesize{$\pm$}0.006 & 0.850\footnotesize{$\pm$}0.007 \\
  DTFD-MIL \cite{Zhang2022DTFDMILDF} &  0.882\footnotesize{$\pm$}0.008 & 0.932\footnotesize{$\pm$}0.016  & 0.816\footnotesize{$\pm$}0.045 & 0.895\footnotesize{$\pm$}0.042  &  0.569\footnotesize{$\pm$}0.026 & 0.847\footnotesize{$\pm$}0.003 \\
  \midrule[0.5pt]
   FT+ CLAM-SB &  0.911\footnotesize{$\pm$}0.017 & 0.956\footnotesize{$\pm$}0.013  & 0.845\footnotesize{$\pm$}0.032 & 0.935\footnotesize{$\pm$}0.027  &  0.718\footnotesize{$\pm$}0.010 & 0.907\footnotesize{$\pm$}0.005 \\

   FT+ TransMIL &  \textbf{0.923\footnotesize{$\pm$}0.012} & \textbf{0.967\footnotesize{$\pm$}0.003}  & \underline{0.848\footnotesize{$\pm$}0.044} & \underline{0.945\footnotesize{$\pm$}0.020}  &  \underline{0.720\footnotesize{$\pm$}0.024} & \underline{0.918\footnotesize{$\pm$}0.004} \\
   FT+ DTFD-MIL &  \underline{0.921\footnotesize{$\pm$}0.007} & \underline{0.962\footnotesize{$\pm$}0.006}  & \textbf{0.849\footnotesize{$\pm$}0.027} & \textbf{0.951\footnotesize{$\pm$}0.016}  & \textbf{0.723\footnotesize{$\pm$}0.008} & \textbf{0.922\footnotesize{$\pm$}0.005} \\
  \midrule[1.2pt]
  Mean-pooling &  0.629\footnotesize{$\pm$}0.029 & 0.591\footnotesize{$\pm$}0.012  &  0.818\footnotesize{$\pm$}0.022 & 0.910\footnotesize{$\pm$}0.032  &  0.350\footnotesize{$\pm$}0.017 & 0.735\footnotesize{$\pm$}0.006 \\
  Max-pooling &  0.805\footnotesize{$\pm$}0.012 & 0.824\footnotesize{$\pm$}0.016  &  0.644\footnotesize{$\pm$}0.179 & 0.826\footnotesize{$\pm$}0.096  &  0.636\footnotesize{$\pm$}0.064 & 0.893\footnotesize{$\pm$}0.019 \\
  KNN (Mean) & 0.468\footnotesize{$\pm$}0.000 & 0.506\footnotesize{$\pm$}0.000 & 0.633\footnotesize{$\pm$}0.066 & 0.749\footnotesize{$\pm$}0.055 & 0.393\footnotesize{$\pm$}0.000 & 0.650\footnotesize{$\pm$}0.000 \\
  KNN (Max) & 0.559\footnotesize{$\pm$}0.000 & 0.535\footnotesize{$\pm$}0.000 & 0.524\footnotesize{$\pm$}0.032 & 0.639\footnotesize{$\pm$}0.063 & 0.477\footnotesize{$\pm$}0.000 & 0.743\footnotesize{$\pm$}0.000 \\
  \midrule[0.5pt]

  FT+ Mean-pooling &  0.842\footnotesize{$\pm$}0.006 & 0.831\footnotesize{$\pm$}0.007  & \textbf{0.866\footnotesize{$\pm$}0.035} & \textbf{0.952\footnotesize{$\pm$}0.018}  &  \underline{0.685\footnotesize{$\pm$}0.014} & \underline{0.900\footnotesize{$\pm$}0.002} \\
   FT+ Max-pooling &  \textbf{0.927\footnotesize{$\pm$}0.011} & \textbf{0.969\footnotesize{$\pm$}0.004}  & \underline{0.852\footnotesize{$\pm$}0.043} & \underline{0.948\footnotesize{$\pm$}0.019}  &  \textbf{0.695\footnotesize{$\pm$}0.013} & \textbf{0.912\footnotesize{$\pm$}0.004}\\
    FT+ KNN (Mean) & 0.505\footnotesize{$\pm$}0.000 & 0.526\footnotesize{$\pm$}0.000 & 0.784\footnotesize{$\pm$}0.044 & 0.907\footnotesize{$\pm$}0.034 & 0.529\footnotesize{$\pm$}0.000 & 0.737\footnotesize{$\pm$}0.000 \\
  
   FT+ KNN (Max)  & \underline{0.905\footnotesize{$\pm$}0.000} & \underline{0.916\footnotesize{$\pm$}0.000} & 0.802\footnotesize{$\pm$}0.063 & 0.882\footnotesize{$\pm$}0.036 & 0.676\footnotesize{$\pm$}0.000 & 0.875\footnotesize{$\pm$}0.000 \\
  % \midrule
  % 0.904  
  \midrule[1.2pt]
  \end{tabular}   
  \caption{
    \textbf{Slide-Level Classification} by using the IN-1K pre-trained backbone or the proposed fine-tuned (FT) in three datasets.
\textbf{Top Rows.} Different MIL architectures are compared to select the top 3 SOTA methods to validate the transfer learning performance using the IN-1K pre-trained backbone or the FT.
\textbf{Bottom Rows.}
The competition of various traditional aggregation and feature evaluation methods by using pre-trained IN-1K or the FT.
  } 
  \vspace*{-5 mm}
  \label{T1} 
  \end{center}   
  \end{table*}

\noindent \textbf{Pretraining and Fine-tuning.}

Our work mainly focuses on the method of fine-tuning. Because good pretrained initialization results in better fine-tuning performance, we employ mainstream pretraining methods: 1) ImageNet-1k (IN-1K) data pretraining. 2) SSL pretraining with SimCLR\cite{chen2020simple}, MoCo\cite{MOCO} and DINO\cite{dino}. 3) for Camelyon-16 and LBP-CECA, since there are comprehensive tumor area annotations, we use the annotation to pretrain the patch backbone, which acts as the upper-bound of our method.
% on which we also compared different annotation ratio's influence because realistic applications generally can annotate limited labels

% 4) More SSL pretraining results is performed and see Appendix e{todo}. Most experiments are mainly performed on IN-1k and SimCLR pretraining, and the difference between all these pretraining methods is assessed in the ablation study.

Different from prior works backbone frozen after pretraining, we perform backbone fine-tuning with the method proposed in 3.4. In this stage,  we fine-tune the backbone and WSI model end-to-end with 25 epochs using the AdamW optimizer, batch size of 1 for WSI with bag size of 512, a learning rate of 1e-5 for backbone and 1e-3 for WSI head. For the BN layers in ResNet, we turn it to eval mode to fix statistics during fine-tuning since we find that the distribution of the top-K instances in a bag is limited to estimate statistics because of the similarity among instances. Shallow layers of backbone are frozen since they only focus on morphological features while deep layer focus on semantics.

\subsection{Slide-level Classification}
\noindent \textbf{Evaluation Metrics.}
For all the experiments, macro-AUC and macro-F1 scores are
reported since all the 3 datasets are class imbalanced.
For Camelyon-16, the official training set is randomly split into training and validation sets with a ratio of 9:1. The experiment is conducted 5 times and the results of the official test set are reported.
For TCGA-BRCA, we perform the 10-folder cross-validation with the same running setting adopted in HIPT \cite{chen2022scaling}.
Besides the dataset LBP-CECA is randomly split with a ratio of 6:1:3 for training, validation, and testing. The experiment is conducted 5 times.
The mean and standard variance values of performance metrics are reported for multi-runs or cross-validation runs.

\noindent \textbf{Comparison with baselines.}
Classification results are summarized in Table \ref{T1}.
We first show full patch supervision results as an upper bound, then we directly evaluate several classic WSI-MIL methods,
including RNN-MIL \cite{campanella2019clinical},
AB-MIL\cite{pmlr-v80-ilse18a}, DS-MIL\cite{li2021dual}, CLAM-SB \cite{lu2021data}, TransMIL \cite{NEURIPS2021_10c272d0}, DTFD-MIL \cite{Zhang2022DTFDMILDF}.
All the WSI-MIL baselines across the three classification tasks suffer from relatively low performance due to the inappropriate backbone features from the pretrained ResNet-50 in IN-1K.
% Then we apply FT features into 3 WSI architectures:
Then we apply FT to the backbone to obtain features for WSI classification by three WSI architectures including
CLAM-SB (pure global attention, no intersection among instances), TransMIL (self-attention for long sequence),
and DTFD-MIL (multi-tier attention paths followed with aggregation, good for patch imbalance).

The results show clearly consistent improvements equipped with FT features under AUC metric.
CLAM-SB achieves a performance relative increase of 9.26\%, 6.37\%, 5.47\% on Camelyon-16, TCGA-BRCA, and LBP-CECA respectively with vanilla FT based on IN-1K pretraining.
% TransMIL achieves a performance increase of 9.51\%, 6.30\%, 8.00\% respectively with vanilla FT.
In the same conditions, TransMIL and DTFD-MIL achieve new SOTAs on all three datasets and can obtain better relative improvement, especially in LBP-CECA with an 8.00\% and 8.85\% growth compared to CLAM-SB.
The results show similar improvements under the F1 metric.
For the bottom two rows of Table \ref{T1},
we compare simple Mean/Max-pooling training and KNN evaluations.

% Mean-pooling achieves a performance increase of 40.61\%, 4.62\%, 22.45\% on Camelyon-16, TCGA-BRCA, and LBP-CECA respectively with vanilla FT based on IN-1k pretraining.
% Max-pooling achieves a performance increase of 14.87\%, 14.77\%, 2.13\% respectively with vanilla FT.
% % While Max-pooling
% KNN(Mean) achieves a performance increase of 3.95\%, 21.09\%, 13.38\% respectively with vanilla FT.
% KNN(Max) achieves a performance increase of 71.22\%, 38.03\%, 17.77\% respectively with vanilla FT.
% The results show similar improvements under the F1 metric.
%

The competitive results demonstrate that our proposed FT method in the backbone of the MIL framework can boost the performance of WSI classification, even for the most simple feature-level Mean/Max-pooling. 
% In addition，......
% The FT features also help to unearth different WSI-MIL architectures' potential,
In addition, task-specific features help us better unearth the properties of different WSI-MIL architectures.
Apparently, the more complex architecture, TransMIL, and DTFD-MIL, can not perform desirable performance with the frozen pretrained parameters in LBP-CECE. In contrast, its capabilities can be further enhanced by fine-tuning the backbone for the defined target.
% Moreover, the effects of different feature aggregation methods varies depending on the task,
Moreover, the effects of different feature aggregation methods vary depending on the task. The better performance is shown by using Max-pooling aggregation in Camelyon-16 containing smaller tumor areas, while Mean-pooling is better in TCGA-BRCA.

\subsection{Combination of SSL and Fine-tuning}
In this section, we further improve the slide-level classification performance by combining SSL with FT. 
\begin{table}[b]
  \begin{center}  
  % \begin{tabular}{*{7}{c}}
  \begin{tabular}{m{2.7cm}<{}||m{2.1cm}<{\centering}m{2.1cm}<{\centering}}
    \midrule[1.2pt]
    %  & \multicolumn{2}{c}{\underline{TCGA-BRCA}}\\
    Method &F1 & AUC \\
    \midrule
  IN-1K $^{\rm \S}$ & - & 0.884\footnotesize{$\pm$}0.059\\
  IN-1K  & 0.797\footnotesize{$\pm$}0.046 &0.879\footnotesize{$\pm$}0.019\\
  /w FT & 0.845\footnotesize{$\pm$}0.032 & 0.935\footnotesize{$\pm$}0.027 \\
  \midrule[0.5pt]
  SimCLR \cite{chen2020simple}$^{\rm \S}$ & -& 0.879\footnotesize{$\pm$}0.069\\
  \midrule[0.5pt]
  MoCo \cite{MOCO} & 0.804\footnotesize{$\pm$}0.042 & 0.904\footnotesize{$\pm$}0.030\\
  % \rowcolor{gray!20}/w FT + top-1k & 0.945  & \\
  /w FT & 0.851\footnotesize{$\pm$}0.029& 0.948\footnotesize{$\pm$}0.026\\
  \midrule[0.5pt]
  DINO \cite{dino} $^{\rm \S}$& -& 0.886\footnotesize{$\pm$}0.059 \\
  DINO & 0.801\footnotesize{$\pm$}0.045 &0.891\footnotesize{$\pm$}0.043\\
  % \rowcolor{gray!20}/w FT + top-1k &  0.963 & \\
  /w FT & 0.848\footnotesize{$\pm$}0.027& 0.944\footnotesize{$\pm$}0.036\\
  \bottomrule[1.2pt]
  \end{tabular}
  \caption{\textbf{Combination of SSL and Fine-tuning.} We compare SSLs with IN-1K and their further improvement via fine-tuning (FT) on TCGA-BRCA. The symbol $^{\rm \S}$ indicates the result released in previous publication \cite{chen2022scaling, chen2022self}.}
  % \vspace*{-5 mm}   
  \label{T2} 
  \end{center}   
  \end{table}

\noindent \textbf{Evaluation Metrics.}
TCGA-BRCA is used for the evaluation of its task complexity in tumor subtyping with the same setting to 4.1.

\noindent \textbf{Comparison with baselines.}
Since there is no apparent single semantic object in small histopathology patches,
we mainly compare SSLs by Contrastive Learning or augmentation like MoCo \cite{MOCO} and DINO \cite{dino}, and SimCLR\cite{chen2020simple}. We show the results of SSLs performed in previous works \cite{chen2022scaling, chen2022self} for fair comparison. All results are performed with the same WSI architecture of CLAM-SB\cite{lu2021data}.

Experimental results are summarized in Table \ref{T2}.
Compared to vanilla IN-1K, MoCo and DINO (combined with FT) achieve consistent growth of 7.85\% and 7.39 \%, respectively.
Besides, they have a slight increase of 1.39\% and 0.96 \% in comparison to IN-1K with FT.
By SSL, the intrinsic task-agnostic features are learned from all patches in WSIs, and after applying our proposed FT for all modules, task-specific features can be distilled from the label and partial data.
In such an SSL with an FT paradigm, the data and the label are explored comprehensively to produce a state-of-the-art WSI analysis.

% We claim that our FT method is orthogonal to SSL.

\subsection{Generalization on Domain Shift}
In this section, we evaluate the generalization of slide-level classification models on domain shift \cite{zhang2022benchmarking} \cite{javed2022rethinking}, which is crucial for real-world clinical applications due to the diversity in staining, preparation, and imaging devices for pathological image processing among hospitals. 

\begin{table}[b]
  \begin{center}  
  % \begin{tabular}{*{7}{c}}
  \begin{tabular}{m{2.5cm}<{}||m{0.9cm}<{\centering}m{0.9cm}<{\centering}|m{0.9cm}<{\centering}m{0.9cm}<{\centering}}
    \midrule[1.2pt]
     & \multicolumn{2}{c|}{\underline{Camelyon-16-C}} & \multicolumn{2}{c}{\underline{Camelyon-17}} \\
    Method &F1&AUC&F1&AUC\\
    \midrule
  Max-pooling &  0.689  &  0.742  & 0.578  & 0.670\\
    % \rowcolor{gray!20}/w FT + top-1k &  0.963 & \\
    /w FT &0.816 & 0.892  &  0.687    &0.720\\
    \midrule[0.5pt]
  
  CLAM-SB \cite{lu2021data}& 0.742  &0.836  & 0.624& 0.702\\
  /w FT & 0.823  & 0.862 &  0.676  & 0.725\\
  \midrule[0.5pt]
  TransMIL \cite{NEURIPS2021_10c272d0} & 0.748 & 0.842 &0.657  & 0.706\\
  /w FT  & 0.795  & 0.857 & 0.684  & 0.717\\
  \midrule[0.5pt]
  DTFD-MIL \cite{Zhang2022DTFDMILDF} & 0.775  & 0.799 & 0.576  & 0.676 \\
  % \rowcolor{gray!20}/w FT + top-1k & 0.945  & \\
  /w FT & 0.804  & 0.838  &0.689  & 0.717\\
  
  \bottomrule[1.2pt]
  \end{tabular}
  \caption{\textbf{Generalization on Domain Shift.} The generalization ability of all methods is compared between fine-tuning(FT) and IN-1K features on two datasets with domain shift. Camelyon-16-C and Camelyon-17 are synthetic and real corruptions respectively.}    
  \label{T3} 
  \end{center}   
  % \vspace*{-5 mm}   
  \end{table}

  \noindent \textbf{Evaluation Metrics.}
  The Camelyon-16-C dataset is generated from Camelyon-16 test set by synthetic domain shift, which adopts a random combination of Brightness, JPEG, and Hue proposed in \cite{zhang2022benchmarking}.
  For Camelyon-17, since it shares similarities to Camelyon-16 but is collected from five different medical centers, we randomly collect 30 samples from each center to evaluate the robustness of natural domain shift.
  We directly evaluate models on Camelyon-16-C and 150 extra data from Camelyon-17 with five times of running same as the previous setting in section 4.1.
  The standard variance values of results is omitted due to the constraint on the paper length and a more clear comprehensive comparison can be found in Supplementary.
  
  \noindent \textbf{Comparison with baselines.}
  The evaluation of generalization results is summarized in Table \ref{T3}. For WSI model architecture, despite CLAM-SB\cite{lu2021data},
  % Generalization evaluation results -> the evaluation of generalization
  TransMIL\cite{NEURIPS2021_10c272d0}, DTFD-MIL\cite{Zhang2022DTFDMILDF}, we also compare Max-pooling on feature level since it shows promising results on Camelyon-16 in 4.1.
  % 回顾一下前面所有的figure ,table 引用 ,要补上和插入一致的额 写法
  % 如 Table.\ref{T1}, Figure.\ref{figxxx}
  
  The results demonstrate consistent improvements after equipping the WSI head with fine-tuned features under the F1 metric:
  In general, all models can resist domain shifts to a certain extent by utilizing the proposed fine-tuning.
  Compared with freezing IN-1K parameters, Max-pooling obtains a noticeable performance increase of 18.43\% and 18.86\% on Camelyon-16-C and Camelyon-17, respectively.
  CLAM-SBachieves an increase of 10.91\%, 8.33\% respectively.
  TransMIL achieves an increase of 6.28\%, 4.11\% respectively.
  % 中间的两个就可以不提了
  DTFD-MIL achieves an increase of 2.90\%, 11.30\% respectively.
  % 最后一个可以说： 相比之下，尽管DTFD本身较强在不同域上的鲁棒性，在我们提出的的FT优化下仍可以进一步提升他的泛化性。
  In contrast, the generalization of DTFD-MIL can still be further improved by our proposed FT optimization, even though it is already robust to different domains.
  
  Interestingly, all improvements in performance under the AUC metric are much less than F1.
  The three Attention-based pooling methods respectively achieve the average improvement of 6.70\% and 7.91\% on Camelyon-16-C and Camelyon-17 under the F1 metric, as compared to 2.93\% and 2.98\% under AUC, which indicates that the classification ability of WSI-MIL model (measured by F1) may be much weaker than ranking (measured by AUC).
  However, doctors may much more care about F1 instead of AUC in clinical diagnosis from sensitivity and specificity perspectives. Most importantly, the classification threshold usually keeps fixed after deploying the model in the real world, which reflects the contribution of our method is more meaningful in practical applications.

\subsection{Further Ablation Experiments}
Additional ablations are included in the Supplementary Materials, with the main focus on the effects of learning rate on the backbone, Number selection of Top-K, and Value selection of Lagrange multiplier.

\subsection{Interpretability and Visualization}
\begin{figure}[htbp]
  \centering
   \includegraphics[width=1.0\linewidth]{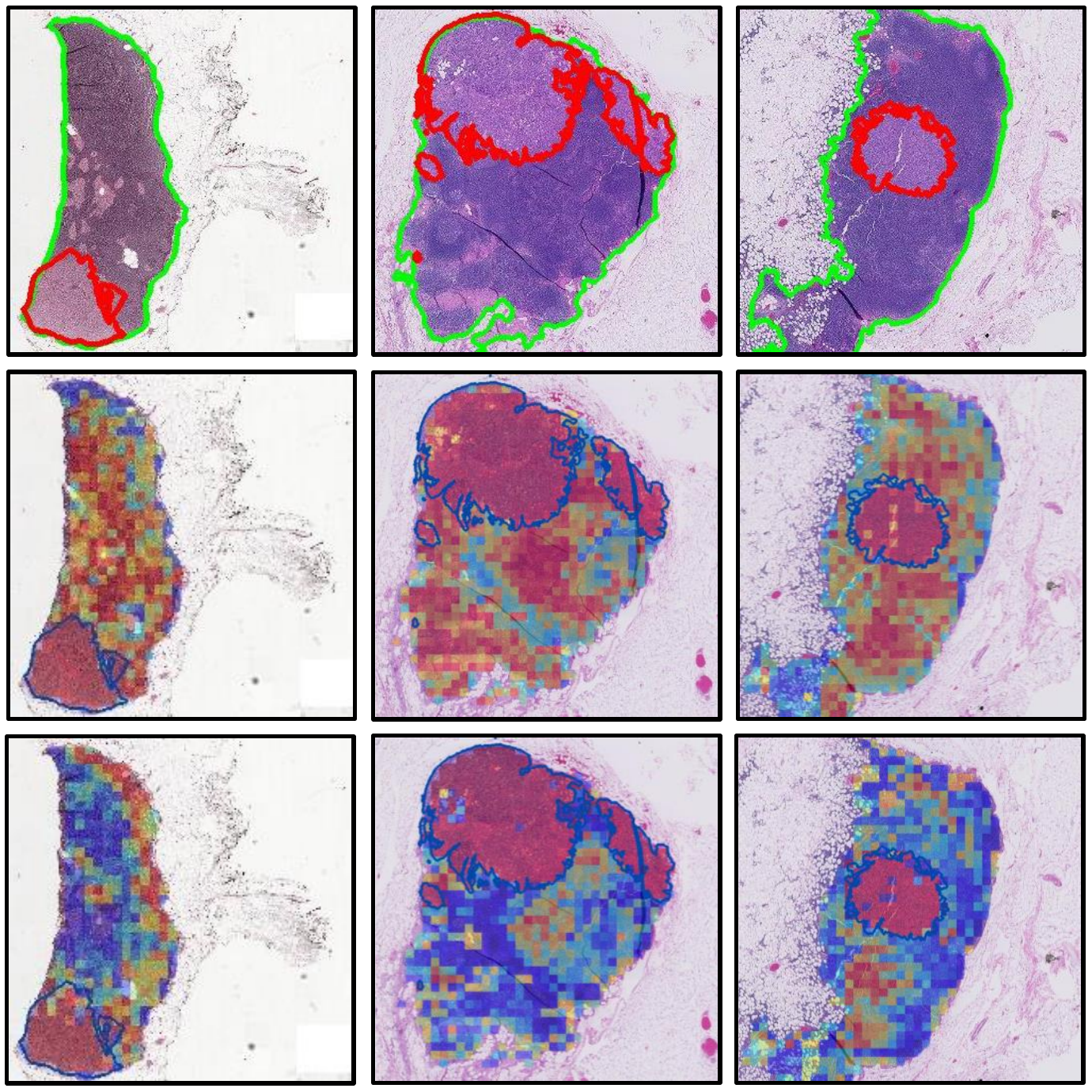}
   \caption{Heatmap comparison between pretraining and FT. 
   The first row shows the full patches label, where red contours denote the tumor area and green contours filter the nonsense background. The second and third row shows the WSI attention map on pretraining and FT features respectively. We can observe that the model is more concentrated on tumors with FT features.}
   \label{FIG4}
   \vspace*{-2.5 mm} 
\end{figure}

Here, we further show the interpretability improvements with FT features. As shown in Figure \ref{FIG4},  attention scores from CLAM-SB\cite{lu2021data} were visualized as a heatmap to determine the ROI and interpret the important morphology used for diagnosis. Obviously, the model is more concentrated on tumors with FT features compared with pretrained.

\section{Conclusion}
In this work, we present a fine-tuning method for WSI classification under the weak supervision of slide-level labels.
Initially, an effective IB module is introduced to mitigate the training cost of Gigapixel WSI, which distills the oversized bag into a sparse one. Then the backbone of the instance is able to be trained end-to-end in the MIL framework by learning and making classifications on the distilled bag. Thus the WSI classification performance is improved through the retention of mainly task-specific information.
In addition, SSL can be combined with the proposed framework for further improvement.
Compared with fully supervised learning, our methods can achieve competitive accuracy by utilizing extremely weak WSI labels.
Furthermore, our training scheme can introduce versatile training-time augmentations for better generalization on datasets with domain shift, which is an inevitable challenge for previous work. The experimental results reflect the advances of our method in both accuracy and generalization.

% Broader impact: % Conclusion 这边一般很少在分出subsection了，
Our proposed approach shows strong potential for MIL application in real-world diagnosis and analysis of pathology images. The proposed feature extraction, with better performance and faster convergence, will be more applicable to real diagnostic WSI learning situations with annotation efficient property.
% We believe that it may bring some inspirations to more diagnosis analysis tasks.

\newpage

%%%%%%%%% REFERENCES
{\small
\bibliographystyle{ieee_fullname}
\bibliography{egbib}
}
\newpage
\begin{appendices}
\part{Supplementary Material}
% \maketitle
\section{Derivation of Variational Information Bottleneck with Bernoulli Prior}
\noindent \textbf{Variational Information Bottleneck} \cite{alemi2017deep}

The Information Bottleneck (IB) can work as an information compression role to intervene in DNN's training \cite{alemi2017deep}. Consider the joint distribution $p(X, Y, Z)$ factors as follows:
\begin{equation}
  \begin{aligned}
p(X, Y, Z) &= p(Z|X, Y )p(Y |X)p(X) \\
&= p(Z|X)p(Y |X)p(X),
  \end{aligned}
\end{equation}
  and assume $p(Z|X, Y ) = p(Z|X)$, corresponding to the Markov chain $Y \leftrightarrow X \leftrightarrow Z$. 
The objective function of IB to be maximized is given in \cite{tishby2000information} as,
  %\begin{center}
  \begin{equation}
  %  \max \{ \underbrace{I(Z, Y)}_{task~specifics} - \overbrace{ \beta I(Z, X)}^{task~irrelevents} \}
  R_{IB} =  I(Z, Y)- \beta I(Z, X),
  \label{EQ3}
   \end{equation}
  where $I(\cdot,\cdot)$ indicates the Mutual Information (MI)
  and $\beta$ is a Lagrange multiplier.
  % and its shares with $X$
  
    % This restriction means that the representation Z cannot depend directly on the labels Y.
    Since the computation of MI is intractable during the training of the neural networks, the variational bound of the two term can be derived as:

    \begin{equation}
        \begin{aligned}
        & I(Z, Y) \\
        &= \int dy dz p(y, z) \log \frac{p(y|z)}{p(y)}
        \\
        &= \int dy dz p(y, z) \log \frac{p(y|z)q(y|z)}{p(y)q(y|z)}
        \\
        &= \int dy dz p(y, z) \{ \log q(y|z) -  \log p(y) + \log \frac{p(y|z)}{q(y|z)} \}
        \\
        &= \int dy dz p(y, z) \log q(y|z) + H(Y) \\ & \qquad + KL(p(Y |Z), q(Y |Z))
        \\&>= \int dy dz p(y, z) \log q(y|z)
        \\&= \int dx dy dz p(x)p(y|x)p(z|x) \log q(y|z),
        % \\ & \approx \frac{1}{N}\sum_{n=0}^N  \mathbb{E}_{z \sim p_{\theta}(z|x_n)}
        % [- \log q_{\phi}(y_n|z)]
        \end{aligned}
    \end{equation}

    \begin{equation}
      \begin{aligned}
      I(Z, X) &= \int dz dx p(x, z) \log \frac{p(z|x)}{p(z)}
      \\ 
      % &= \int dz dx p(x, z) \log p(z|x) - \int dz dx p(z) \log p(z)
      % \\
      &= \int dz dx p(x, z) \log \frac{p(z|x)r(z)}{p(z)r(z)}
      \\ 
      &= \int dz dx p(x, z) \log \frac{p(z|x)}{r(z)} - KL(p(Z), r(Z))
      % \\ 
      % &= \int dy dz p(y, z) \log q(y|z) + H(Y) \\ & \qquad + KL(p(Y |Z), q(Y |Z))
      \\&<= \int dz dx p(x, z) \log \frac{p(z|x)}{r(z)}
      \\&= \int dz dx p(x) p(z|x) \log \frac{p(z|x)}{r(z)},
      \end{aligned}
  \end{equation}

    Thus, the IB objective can be transferred as a variational bound of Eq.\eqref{EQ3} as follows:
  \begin{equation}
    \begin{aligned}
  R_{IB} &>= \int dx dy dz p(x)p(y|x)p(z|x) \log q(y|z) 
  \\& - \beta \int dz dx p(x) p(z|x) \log \frac{p(z|x)}{r(z)} 
  \\&=
  -\frac{1}{N}\sum_{n=0}^N  \mathbb{E}_{z \sim p_{\theta}(z|x_n)}
        [- \log q_{\phi}(y_n|z)] -
    \\ & \beta KL[p_{\theta}(z|x_n),r(z)],
    \end{aligned}
    \label{EQx}
  \end{equation}
  Where the $p(x)p(y|x)$ is approximated by using the empirical data distribution during stocastic batch iteration training, $N$ denotes the number of samples,
  $q_{\phi}(y|z)$ is a parametric approximation to the likelihood $p(y|z)$,
  %用逗号不用分号
  $r(z)$ is the prior probability of $z$ to variational approximate the marginal $p(z)$, and $p_{\theta}(z|x)$ is the parametric posterior distribution over $z$.
  Then, to maximize IB objective can be seen to minimize:
  \begin{equation}
    \begin{aligned}
  J_{IB} = \frac{1}{N}\sum_{n=0}^N  \mathbb{E}_{z \sim p_{\theta}(z|x_n)}
        [- \log q_{\phi}(y_n|z)] +
    \\ \beta KL[p_{\theta}(z|x_n),r(z)].
    \end{aligned}
    \label{EQ4}
  \end{equation}

  % where $n, N$ denote the sample's index and quantity respectively;

  %需要起个头

  \noindent \textbf{Learn Sparsity via Variational Bound of IB \cite{paranjape-etal-2020-information}}

  To trade off the dilemma of computational limitation and task-specific representation learning via end-to-end back-propagation, we propose to utilize the IB module to filter most task-irrelevant instances for task-specific fine-tuning.

  The above filtering process can be implemented by optimizing the second term of in Eq.\eqref{EQ3} which controls the compression. There are two ways that compress $X$ to $Z$ by decreasing the KL divergence between $p(z|x)$ and $r(z)$ in Eq.\eqref{EQ4} variational method: reducing the dimension of representation $Z$ compared to $X$ in \cite{alemi2017deep}, or converting input $X$ into a sparse one in \cite{paranjape-etal-2020-information}.

  For the setting of our long instance sequenced MIL, we reduce $I(X, Z)$ into a degree so that the gradients can be back-propagated to the backbone encoder, which needs us to convert a WSI of bag size over 10k into 1k for the sake of sparsity. Considering MIL for tumor v.s. normal binary classification without loss of generality and the latent label ${y_i}$ of each instance $x_i$, we argue that it is sufficient enough to make the WSI level prediction if one tumor area is detected.
  With the above understanding, we propose to learn compressed components similar to \cite{paranjape-etal-2020-information} by defining a IB module as:
  \begin{equation}
  z = m \odot x,
  \label{EQ5}
  \end{equation}
  
  where $m$ is a Bernoulli$(\pi)$ distributed binary mask, thus $r(z|x) = (1-\pi)\delta(z) + \pi\delta(z - x)$. and in this way $KL[p_{\theta}(z|x),r(z)]$ in Eq.\eqref{EQ4} can be decomposed as,
  \begin{equation}
  \begin{aligned}
    & KL[p_{\theta}(z|x),r(z)] 
    \\& = (1-\theta(x)) \int \delta(z) log \frac{p_{\theta}(z|x)}{r(z)}dz 
    \\ &+ \theta(x) \int \delta(z-x) log \frac{p_{\theta}(z|x)}{r(z)}dz 
    % \\& = \int dz dx p(x) p(z|x) \log \frac{p(z|x)}{r(z)}
    % \\ & = \int dz dx p(x) p(z|x) \log \frac{p(z|x)}{r(z)}
    \\& = (1-\theta(x)) log \frac{1- \theta(x)}{1-\pi} + \theta(x) log \frac{\theta(x)}{\pi p(x)}
    \\ & = KL[p_{\theta}(m|x),r(m)] - \theta(x) log p(X)
    \\ & = KL[p_{\theta}(m|x),r(m)] + \pi H(X),
    \end{aligned}
  \end{equation}
  % where $H(X)$ is the entropy of $X$,
  % thus it can be dropped since it is constant.
  where $H(X)$ is the entropy of $X$, which can be omitted during the minimization due to its constant value. 

\section{Connections to Sparse Attention}
Coming soon.

\section{PyTorch Pseudocode}
We show the pytorch pseudocode of the WSI sparsity training of stage-1.
\begin{algorithm}[h]
  \SetAlgoLined
  \PyComment{Learn sparsity of WSI with fixed backbone} \\
  for (X,y) in data\_loader:\\
  \Indp   % start indent
    with torch.no\_grad():\\
  \Indp   % start indent
      model.eval()\\
      Z\_0 = model(X)\\
      \PyComment{X = {x\_1,x\_2,...,x\_n}} \\
      \PyComment{Z = {z\_1,z\_2,...,z\_n}} \\
  \Indm
      model.train()\\
      \PyComment{IB is a sequential FCs} \\
      M = IB(Z\_0)\\
      logits = torch.sigmoid(M)\\
      p\_z = Bernoulli(logits)\\
      Z\_mask = p\_z.sample()\\
      r\_z = Bernoulli($\pi$)\\
      \PyComment{reparameterization trick for Bernoulli samples} \\
      Z\_1 = Z\_0$\cdot$(M+Z\_mask)/2\\ 
      Y = model\_wsi(Z\_1)\\
      loss1 = CrossEntropyLoss(Y,y)\\
      loss2 = KL\_divergence(p\_z, r\_z)\\
      loss = loss1+$\beta$loss2 \\
      optimizer.zero\_grad()\\
      loss.backward() \\
      optimizer.step() \\
 
  \caption{PyTorch-style pseudocode for WSI task-specific IB sparsity learning}
  \label{alg}
 \end{algorithm}

\section{Details of Datasets}
\noindent \textbf{Camelyon-16} \cite{bejnordi2017diagnostic} is a public dataset for metastasis detection in breast cancer (tumor / normal classification), including 270 training sets and 130 test sets. A total of about 1.5 million patches at ×20
magnification are obtained after pre-process.

\noindent \textbf{TCGA-BRCA}
The Cancer Genome Atlas Breast Cancer \cite{petrick2021spie} is a public dataset for breast invasive carcinoma cohort for  Invasive Ductal Carcinoma (IDC) versus Invasive Lobular Carcinoma (ILC)
subtyping. The WSI is segmented into non-overlapping tissue-containing patches at 20× magnification and about 2.0 million patches were curated from 1038 WSIs.

\noindent \textbf{LBP-CECA}
The Liquid-based Preparation cytology
for Cervical Cancer’s early lesion screening dataset is introduced to
validate the universality of our method on cyto-pathology. The WSIs include 4 classes (Negative, ASC-US, LSIL, ASC-H/HSIL \cite{nayar2015bethesda}) and are segmented into patches with  overlapping of 25 and size of 256 at 20× magnification and about 3.2 million patches were curated from 1393 WSIs.

\noindent \textbf{Camelyon-16-C} is generated with random synthetic domain
shift on Camelyon-16\cite{bejnordi2017diagnostic} testset for simulation. 
Three kind of corruptions are included: Jpeg compression, Brightness and Hue are implemented by the code in \cite{zhang2022benchmarking}, all with a severity of 2.

\noindent \textbf{Camelyon-17} \cite{litjens20181399} dataset is collected from five different centers. It is an offical extension challenge of Camelyon-16. In this paper we combine all tumor positive WSI and random selected negative to constitude a real domain shift test set. Finally, 164 WSIs are sampled out for test.

% \title{Supplementary Material}
% \maketitle

\section{Further Ablation Experiments}
\noindent \textbf{Influence of Learning Rate on the Backbone}

Here we show the influence of backbone learning rate on Top-512 fine-tuning results, which is performed on Camelyon-16 only once for the relatively long training time of training stage-2. The ablations results are summarized in Table \ref{T1}. Since the supervision signal of WSI is too weak, we find that lower learning rate helps convergence. For learning rate of 1e-3 and 5e-4, the fine-tuning collapse quickly and diverges to Nan loss. For learning rate of 1e-5, we get the best fine-tuning results on Top-512 as a WSI distilled bag.

\begin{table}[htbp]
  \begin{center}  
  % \begin{tabular}{*{7}{c}}
  \begin{tabular}{m{1.0cm}<{}||m{1.5cm}<{\centering}m{1.5cm}<{\centering}}
    \midrule[1.2pt]
    %  & \multicolumn{2}{c}{\underline{TCGA-BRCA}}\\
    LR & F1 & AUC \\
    \midrule
   1e-3 & N/A & N/A \\
   5e-4 & N/A & N/A  \\
   1e-4 &  0.682 & 0.744  \\
   5e-5 &  0.713 & 0.741 \\
   1e-5 &  \textbf{0.899} & \textbf{0.944} \\
   5e-6 &  0.876 & 0.908 \\
   1e-6 &  0.806 & 0.804 \\
  \bottomrule[1.2pt]
  \end{tabular}
  \caption{\textbf{Influence of Learning Rate on the Backbone} during fine-tuning process with weakly WSI supervision. } 
  % \vspace*{-5 mm}   
  \label{T1} 
  \end{center}   
  \end{table}

\noindent \textbf{Number selection of Top-K}

Here we show the influence of IB module training in stage-1, which is performed on Camelyon-16 with five runs. The ablations results are summarized in Table \ref{T2}. Generally, with the increasement of K, less essential instances would be neglected, resulting in better performace.
However, most of WSIs in the Camelyon-16 dataset are with only a few tumor area, thus the less Top-K somehow fit better this dataset property. So we find that top-2048 shows the best results and even higher than all instances used for WSI decision. However for the computational limitation, we finally select top-512 for fine-tuning of stage-2. 

\begin{table}[htbp]
  \begin{center}  
  % \begin{tabular}{*{7}{c}}
  \begin{tabular}{m{2.0cm}<{}||m{2.1cm}<{\centering}m{2.1cm}<{\centering}}
    \midrule[1.2pt]
    %  & \multicolumn{2}{c}{\underline{TCGA-BRCA}}\\
    Top-K & F1 & AUC \\
    \midrule
   128 & 0.840\footnotesize{$\pm$}0.011 & 0.870\footnotesize{$\pm$}0.010  \\
   256 &  0.843\footnotesize{$\pm$}0.009 & 0.870\footnotesize{$\pm$}0.010   \\
   512 &  0.843\footnotesize{$\pm$}0.005 & 0.866\footnotesize{$\pm$}0.011  \\
   1024 &  0.845\footnotesize{$\pm$}0.007 & 0.864\footnotesize{$\pm$}0.011  \\
   2048 &  \textbf{0.846\footnotesize{$\pm$}0.004} & \textbf{0.875\footnotesize{$\pm$}0.010}  \\
   \midrule
    all & 0.839\footnotesize{$\pm$}0.018 & 0.875\footnotesize{$\pm$}0.028 \\
  \bottomrule[1.2pt]
  \end{tabular}
  \caption{\textbf{Number selection of Top-K.} } 
  % \vspace*{-5 mm}   
  \label{T2} 
  \end{center}   
  \end{table}

\noindent \textbf{Value selection of Lagrange multiplier}

Here we show the influence of Lagrange multiplier during training stage-1, which is performed on Camelyon-16 with five runs. Definitely, the Lagrange multiplier $\beta$ works as a trade off factor of the two task: if we care more about WSI training loss with a low $\beta$, then the ranking or sparsity properties of IB module may not be well learned. On the contrary, a large $\beta$ will influence the training of WSI classifier. The ablations results are summarized in Table \ref{T3} and we find that the best selection of $\beta$ is 1e-1. 

\begin{table}[htbp]
  \begin{center}  
  % \begin{tabular}{*{7}{c}}
  \begin{tabular}{m{2.0cm}<{}||m{2.1cm}<{\centering}m{2.1cm}<{\centering}}
    \midrule[1.2pt]
    %  & \multicolumn{2}{c}{\underline{TCGA-BRCA}}\\
    $\beta$ & F1 & AUC \\
    \midrule
    Upper bound & 0.839\footnotesize{$\pm$}0.018 & 0.875\footnotesize{$\pm$}0.028 \\
    \midrule
    1e-3 & 0.835\footnotesize{$\pm$}0.008 & 0.860\footnotesize{$\pm$}0.012  \\
   1e-2 & 0.833\footnotesize{$\pm$}0.006 & 0.860\footnotesize{$\pm$}0.028  \\
   1e-1 &  \textbf{0.849\footnotesize{$\pm$}0.010} & \textbf{0.865\footnotesize{$\pm$}0.014}    \\
   1 &  0.839\footnotesize{$\pm$}0.015 & 0.852\footnotesize{$\pm$}0.018    \\
   10 &  0.838\footnotesize{$\pm$}0.016 & 0.862\footnotesize{$\pm$}0.020    \\
   100 &  0.828\footnotesize{$\pm$}0.010 & 0.853\footnotesize{$\pm$}0.007    \\
  \bottomrule[1.2pt]
  \end{tabular}
  \caption{\textbf{Value selection of Lagrange multiplier.} } 
  % \vspace*{-5 mm}   
  \label{T3} 
  \end{center}   
  \end{table}

\section{Result Analysis of the 3 Stages}
\begin{figure}[htbp]
    \centering
	
	\begin{minipage}{0.45\linewidth}
		\centering
        % \vspace*{-3 mm}
         \caption{Performance of three stages on Camelyon-16, most can be found from the prior submission material.} 
         \begin{tabular}{{l|c}}
  % \vspace*{-8 mm}
    \midrule[1.0pt]
       % & \multicolumn{2}{c}{\underline{Camelyon-16}} \\
        Method & AUC\\
          % 0.849±0.010 0.865±0.014
          \midrule
        
          CLAM-SB &  0.875\\
          stage-1 &  0.865 \\
          stage-2 &  0.944 \\
          stage-3 &  0.956 \\
          stage-2 random  &  0.731 \\
          \midrule[1.0pt]
          \end{tabular}

          % \vspace*{-8 mm}
          \label{t5} 
	    \end{minipage}
        \vspace*{-2 mm}
\hfill
     \begin{minipage}{0.45\linewidth}
		\centering
   % \vspace*{-3 mm}
        \caption{A WSI with very few tumor areas (blue).}
		% \vspace{-0.6cm}
		\setlength{\abovecaptionskip}{0.28cm}
		\includegraphics[width=\linewidth]{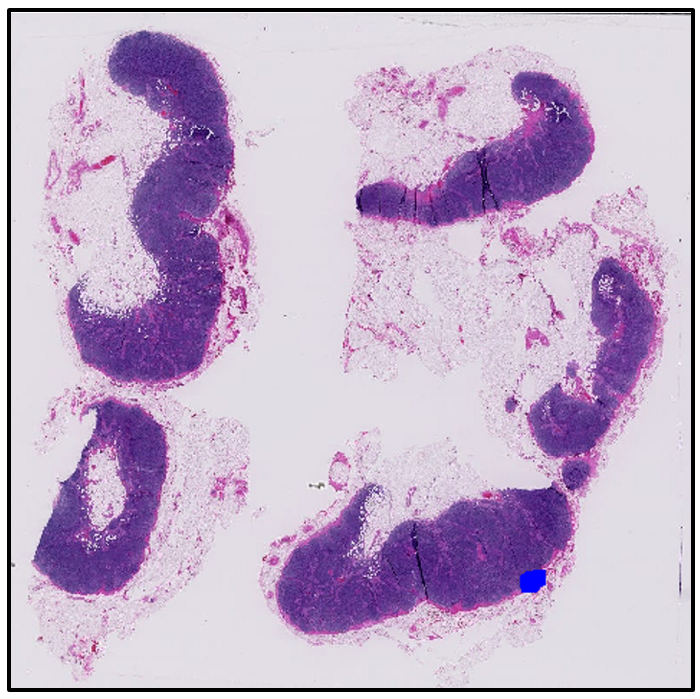}
		
		\label{f5}
	\end{minipage}
        \vspace*{-2 mm}
	%\qquad
	
\end{figure}

There is a probability that the top-K instances may not contain at least one tumor patch for extreme cases, e.g. some Camelyon-16 WSIs contain very few tumors in Fig.\ref{f5}. Thus stage-3 is needed for covering all instances to get WSI result equipped with fine-tuned backbone, which shows further improvement compared to stage-2 in Fig.\ref{t5}. 
We also show that with random k instances, the model in stage-2 cannot converge, in Fig.\ref{t5}.
  
    \end{appendices}
\end{document}